# Open source software for automatic subregional assessment of knee cartilage degradation using quantitative T2 relaxometry and deep learning

Kevin A. Thomas, BSE[1] • Dominik Krzemiński, MS[2] • Łukasz Kidziński, PhD[3] • Rohan Paul, MS[1] • Elka B. Rubin, BS[4] • Eni Halilaj, PhD[5] • Marianne S. Black, PhD[4] • Akshay Chaudhari, PhD[1,4] • Garry E. Gold, MD[3,4,6] • Scott L. Delp, PhD[3,6,7]

(1) Department of Biomedical Data Science, Stanford University, California, USA (2) Cardiff University Brain Research Imaging Centre, Cardiff University, United Kingdom (3) Department of Biomedical Engineering, Stanford University, California, USA (4) Department of Radiology, Stanford University, California, USA (5) Department of Mechanical Engineering, Carnegie Mellon University, Pennsylvania, USA (6) Department of Orthopaedic Surgery, Stanford University, California, USA (7) Department of Mechanical Engineering, Stanford University, California, USA

Correspondence: Kevin A. Thomas (email: *kevin.a.thomas@stanford.edu*)

## Abstract

**Objective**: We evaluate a fully-automated femoral cartilage segmentation model for measuring T2 relaxation values and longitudinal changes using multi-echo spin echo (MESE) MRI. We have open sourced this model and corresponding segmentations. **Methods**: We trained a neural network to segment femoral cartilage from MESE MRIs. Cartilage was divided into 12 subregions along medial-lateral, superficial-deep, and anterior-central-posterior boundaries. Subregional T2 values and four-year changes were calculated using a musculoskeletal radiologist's segmentations (Reader 1) and the model's segmentations. These were compared using 28 held out images. A subset of 14 images were also evaluated by a second expert (Reader 2) for comparison. **Results**: Model segmentations agreed with Reader 1 segmentations with a Dice score of $0.85 \pm 0.03$. The model's estimated T2 values for individual subregions agreed with those of Reader 1 with an average Spearman correlation of 0.89 and average mean absolute error (MAE) of 1.34 ms. The model's estimated four-year change in T2 for individual regions agreed with Reader 1 with an average correlation of 0.80 and average MAE of 1.72 ms. The model agreed with Reader 1 at least as closely as Reader 2 agreed with Reader 1 in terms of Dice score (0.85 vs 0.75) and subregional T2 values. **Conclusions**: We present a fast, fully-automated model for segmentation of MESE MRIs. Assessments of cartilage health using its segmentations agree with those of an expert as closely as experts agree with one another. This has the potential to accelerate osteoarthritis research.

# 1     Introduction

Knee osteoarthritis (OA) affects 20-40% of the U.S. population over 65 years of age and currently has no cure[1–3]. Diagnosis and measurement of patients' OA severity typically relies on the use of X-rays, which only enable the detection and assessment of OA that has progressed to the point of joint space narrowing and visible changes to the bone[4,5]. Novel quantitative magnetic resonance imaging (qMRI) techniques have the potential to measure early changes in cartilage matrix composition before the onset of gross structural changes[6,7]. For example, T2 relaxation time mapping from multi-echo spin-echo (MESE) T2-weighted MRIs potentially reflects the hydration level and collagen of the matrix[8]. As OA progresses, T2 values in affected cartilage tend to increase[9]. Because the disease takes years to progress at the structural scale, T2 relaxometry potentially offers the ability to measure the effect of therapeutic interventions earlier than is possible with radiographs or structural MRI[10]. Early measurement of cartilage changes may also enhance treatment efficacy by enabling earlier intervention.

Currently, manual or semi-automated segmentation of cartilage in MRI scans is a necessary first step for assessing cartilage health from these images. Most work on automatic segmentation of cartilage in MRIs has focused on structural sequences used for morphological assessments (e.g. thickness and volume)[11–16]. While these sequences feature higher cartilage-to-background contrast and higher resolution than qMRI methods, the segmentation of MESE MRIs is crucial for the assessment of early changes in cartilage composition.

Prior studies have used image registration to transfer the segmentation of morphological images to corresponding MESE images of the same patient taken in succession[17–19]. However, registration can be imperfect, partially due to the potential for non-affine movement of the knee throughout the acquisition time of the two images. The risk of patient movement and registration error may be higher when the time between the acquisition of the morphological image and T2 image is longer. For example, the image acquisition protocol for the Osteoarthritis Initiative (OAI), a longitudinal study of OA, separated acquisition of the morphological images from the MESE images by 18 minutes[20]. Significant differences in the contrast and resolution between the morphological images and MESE images in the OAI also contribute to imperfect registration.

Researchers often manually segment MESE images[21–23]. This choice may be influenced by the potential errors of image registration-based segmentation and the lack of time and resources to acquire the additional morphological images necessary for current automated segmentation approaches. However, manual segmentation is time-intensive[24]. This bottleneck leads researchers to restrict their analyses to small subregions of cartilage or individual MRI slices[25–30], losing information on the spatial variability of cartilage health.

Deep learning is valuable for automated medical image analysis. It allows for a mapping from an input (e.g. MRI image) to an output (e.g. segmentation) to be learned from example data (e.g. MRIs previously segmented by an expert)[12,31]. Two central components of deep learning models are architecture and parameters. A model's architecture determines the strategy of the mapping, while a model's parameters determine the specific calculations used to obtain the output given the input[32]. The parameters' values are tuned to their final values using the example data, a process known as model training. Training deep learning models is computationally expensive but using a trained model to segment a new image is fast and computationally cheap. This differs from other segmentation strategies, like atlas-based approaches, which require iterative, computationally expensive deformations of one or more segmented template images to segment each new image.

A few recent studies have aimed to automate the segmentation of MESE images directly[31,33]. These have used atlas-based approaches or a combination of deep learning with simplex deformable modeling. While they have reported promising results, the ability of fully automated methods to produce accurate, subregional measurements of cross-sectional and longitudinal variations in T2 has not been explored.

The aim of this work was to develop a fully automated femoral cartilage segmentation model that operates directly on MESE images and to evaluate the model as a means to measure individuals' subregional T2 values longitudinally. A model that can take a MESE MRI as input and produce expert-quality assessments of T2-based cartilage health and disease progression would enable more accurate, efficient OA research.

## 2    Methods

### 2.1    Overview

We used 286 MESE MRI volumes from 143 subjects from the OAI. Each MRI was segmented with a semi-automated process and refined by a radiologist. These segmentations were used as ground truth. We used a Convolutional Neural Network (CNN) to learn MRI features predictive of cartilage location. The model was trained using training set images and their ground-truth segmentations, tuned using a validation set, then evaluated using a test set. Agreement between predicted segmentations and ground-truth segmentations was assessed using the Dice score and Jaccard index. Segmented cartilage was divided into 12 subregions. The average T2 value and four-year change in T2 value was calculated for each subregion of each subject using human segmentation and the CNN segmentation. Focal areas of increased T2 value were identified automatically and the percentage of total cartilage area covered by these lesions was compared between segmentation approaches.

### 2.2    Data

The OAI is a public study of knee OA in which MRIs were collected longitudinally. We used 286 sagittal plane MESE MRI volumes (Table 1) from 143 OAI subjects assessed at baseline and four years later. Half of the subjects were randomly selected from among those in the OAI Incidence Cohort with BMI > 30 and the other half were age- and sex-matched controls with normal BMI in the Incidence Cohort. All had a Kellgren-Lawrence grade of 0 in the imaged knee, indicating no radiographic OA, both at baseline and four years later. They were determined to be at risk for developing OA as determined by knee symptoms and at least two other risk factors (e.g. family history, previous knee injury, occupational burden). Subjects were 48% female, 90% Caucasian/9% Black or African American/1% Asian, aged 45-78 years, and had BMIs of 18.4-44.3 kg/m$^2$. Each image was segmented with a semi-automated process and refined by an experienced musculoskeletal radiologist (Qmetrics Technologies, Pittsford, New York), referred to as Reader 1. Subjects were split into training (115 subjects, 230 image volumes), validation (14 subjects, 28 image volumes), and test (14 subjects, 28 image volumes) sets with no crossover of subjects across sets. The subjects' OAI patient identification numbers and train/validation/test set assignments are available in Supplementary Material 1.

### 2.3    Image Processing

The model was designed to take in the second echo of a slice to produce that slice's segmentation. However, while training the network, a given slice's first echo was used with 20% probability, the second echo was used with 60% probability, and the third echo was used with 20% probability. This echo selection process was performed for each slice in each epoch (i.e., each round of training in which a model sees every training image once). Exposing the model to a range of decay times was done to improve the model's performance on new patients whose T2 values and second echoes may vary from those in the training set. To increase the model's sensitivity, 90% of slices that did not contain cartilage were randomly removed from each image volume in each epoch. Only the second echo was used when evaluating models with the validation and test sets and all slices of these images were retained.

Before an echo image was used as input to the model, its voxel values were shifted and rescaled so its median voxel value was 0, its 25th percentile voxel value was -1 and its 75th percentile voxel value was 1. Voxel values were trimmed between 3rd and 97th percentile to remove outliers.

Table 1: OAI MESE MRI imaging parameters

| Parameter | Value |
|---|---|
| Matrix (phase) | 269 |
| Matrix (frequency) | 384 |
| Number of slices | 21 |
| FOV (mm) | 120 |
| Slice thickness/gap (mm/mm) | 3 / 0.5 |
| TE (ms) | 10, 20, 30, 40, 50, 60, 70 |
| TR (ms) | 2700 |
| X-resolution (mm) | 0.313 |
| Y-resolution (mm) | 0.446 |

## 2.4 Segmentation Model Training

We used the two-dimensional U-Net[34], a CNN architecture shown to perform well on medical image segmentation tasks, including cartilage segmentation in morphological MRIs[12]. We trained the CNN to learn MRI features that were predictive of cartilage location from the dataset. It was optimized to produce segmentations with high Dice scores relative to the ground truth segmentations (Supplementary Material 2: Loss function). The architecture was designed to segment each slice independently.

We trained a variety of models, all with the U-Net architecture, but different hyperparameters (Supplementary Material 3: Training hyperparameters). Each model was evaluated with the validation set images. The model with the highest average validation set Dice score was selected as the final model and was then evaluated with the test set. This enabled the final model's performance on the test set to serve as an objective measure of how it would perform on new images not used in this study. See Supplementary Material 4 for specifications of the hardware and software used.

## 2.5 Segmentation Refinement

For each voxel of an input image slice, the CNN outputs a value, $p \in [0, 1]$. Larger values in this range can be interpreted as voxels that the model predicts to contain cartilage. All voxels with $p > .01$ were initially considered as potentially containing cartilage. T2 values were calculated for each of these voxels for both expert and model segmentations. The first echo was excluded to minimize stimulated-echo artifacts[35] and a noise-corrected monoexponential fit was used. Segmented voxels with a T2 value outside the physiological range of cartilage, [0,100] ms, were discarded. This refinement procedure was done for both ground truth segmentations and the model's segmentations.

In early experiments, models frequently predicted small amounts of cartilage at the medial and lateral joint margins in slices that did not contain cartilage. To reduce this, a threshold was used to set a minimum number of cartilage voxels per slice. In slices that were predicted to have fewer cartilage voxels than the threshold, all of these voxels were discarded from the model's segmentation. Another threshold was applied to $p$ such that any $p \geq threshold$ was considered cartilage and all other outputs were not. The validation image set was used to set the values of these two thresholds to maximize the Dice score between model segmentations and Reader 1 segmentations. These thresholds, identified to be 425 voxels and $p \geq 0.501$ via grid search, were applied to all test set segmentations to binarize them.

## 2.6 Segmentation Evaluation
### 2.6.1 Direct comparison on segmentation masks

To compare the agreement between the model and Reader 1, the volumetric Dice score and volumetric Jaccard index were calculated for each image's segmentations. To understand how this compares with inter-reader agreement, an image volume from each test set subject, 14 image volumes total, was manually segmented by a researcher with extensive cartilage segmentation experience (ER, referred to as Reader 2) for comparison with the ground truth segmentations. These segmentations were reviewed by a musculoskeletal subspecialty radiologist with several years of experience in MESE knee MRI and who had previously served on the Imaging Advisory Board for the OAI (GER).

### 2.6.2 Comparison of subregional T2 mean values

To investigate the model's impact on T2 relaxometry, we used it to evaluate the average T2 value and four-year change in average T2 value in anatomical subregions of the femoral cartilage delineated along medial-lateral, superficial-deep, and anterior-central-posterior boundaries. These values were then compared to the values obtained via Reader 1 segmentations. Anatomical subregions were obtained by first projecting the cartilage onto a two-dimensional plane using previously validated techniques[36], then dividing it into 12 subregions automatically (Figure 4).

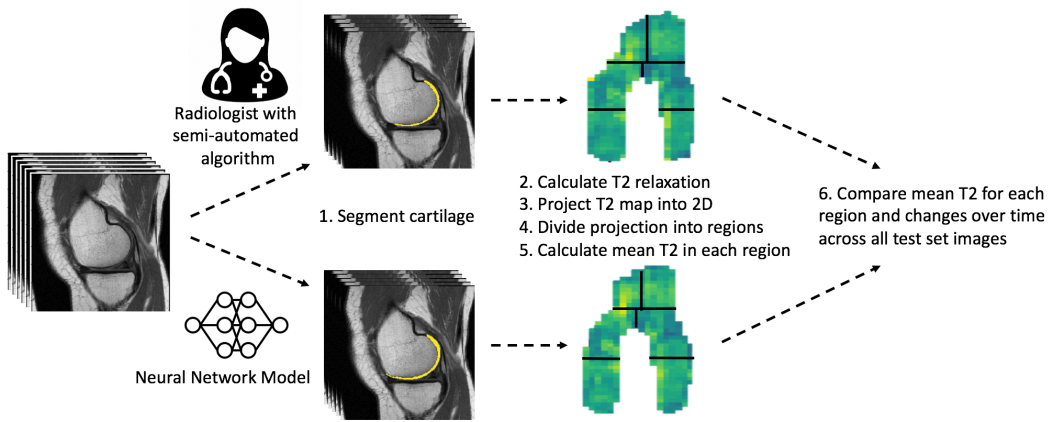

**Figure 1: Segmentation evaluation procedure.** Each test set image was segmented by both Reader 1 and the model. The T2 value for segmented voxels was calculated for each segmentation and the resulting T2 maps were projected into 2D. The projected T2 maps were divided into anatomical subregions and the mean T2 in each subregion was compared, as well as the change in mean T2 in each subregion over time.

For each subregion, the Spearman correlation and mean absolute error (MAE) between the model's estimated values and the values derived from Reader 1 were assessed. Reader 2 segmentations were evaluated in the same way, via comparison to Reader 1 segmentations, and these results were compared with the model's results.

### 2.6.3 Comparison of focal T2 elevation

Cluster analysis was performed to identify focal subregions of elevated T2, according to previously validated approaches[36]. T2 maps for the 3D cartilage were calculated for each image, then projected into 2D. The projection maps for the two imaging timepoints of a subject were registered and subtracted to identify the change in T2 value at each pixel location over time. Clusters of contiguous pixels that were all more than one standard deviation above the subject's mean T2 change across the full cartilage plate were noted. Clusters that covered more than 1% of the area of the cartilage plate were identified and labeled as focal lesions. The cluster intensity and area thresholds remove noise but still identify focal defects. The area of a subject's femoral cartilage plate covered by clusters was calculated. This has been proposed as a measure of OA risk and progression.

# 3 Results

## 3.1 Training and evaluation time

The final model was trained in 17 epochs, requiring 28 hours. When using the model to segment a new image volume, it required approximately 6 seconds using an NVIDIA K80 graphical processing unit.

## 3.2 Direct comparison of segmentation masks

With Reader 1 segmentations used as ground truth, the model had an average volumetric Dice score of $0.85 \pm 0.03$ and an average volumetric Jaccard index of $0.74 \pm 0.04$ for the full test set (Table 2). For the test subset segmented by Reader 2, the model had an average volumetric Dice score of $0.85 \pm 0.03$ and an average volumetric Jaccard index of $0.73 \pm 0.05$ with respect to Reader 1, while Reader 2 had an average volumetric Dice score of $0.74 \pm 0.03$ and an average volumetric Jaccard index of $0.59 \pm 0.04$ with respect to Reader 1. With Reader 2 held as ground truth, the model had an average volumetric Dice score of $0.75 \pm 0.03$ and an average volumetric Jaccard index of $0.60 \pm 0.04$.

Table 2: Segmentation comparison between readers and model

|  | **Model vs Reader 1 (Full Test Set)** | **Model vs Reader 1 (14 MRI test subset)** | **Reader 1 vs Reader 2 (14 MRI test subset)** | **Model vs Reader 2 (14 MRI test subset)** |
|---|---|---|---|---|
| **Dice Score** | $0.851 \pm 0.029$ | $0.845 \pm 0.031$ | $0.741 \pm 0.030$ | $0.753 \pm 0.027$ |
| **Jaccard Index** | $0.742 \pm 0.043$ | $0.732 \pm 0.046$ | $0.590 \pm 0.037$ | $0.605 \pm 0.035$ |

Volumetric Dice scores and volumetric Jaccard indices were calculated for each test set image. Values reported here are the mean value ± standard deviation calculated across the 28 image volumes in the test set or 14 image volumes in the test subset.

## 3.3 Comparison of subregional T2 mean values

The model's estimates of subregional average T2 values were significantly correlated with those of Reader 1 for all subregions ($p <$ 1e-5). Spearman correlations and the MAE of the model's estimates of subregional average T2 are shown in (Figure 4, Left). Ten of the 12 smallest subregions' model estimates did not have significant bias ($p > 0.05$) while the superficial lateral anterior subregion had a bias of 0.869 ms and the deep medial central subregion had a bias of -1.02 ms (Supplementary Material 6: Bland Altman Plots). Bland Altman plots for the full deep region and full superficial region are shown in Figure 2.

For the images segmented by Reader 2, subregional average T2 values derived from the model's segmentations differed from those of Reader 1 with similar magnitude as did the subregional average T2 values derived from Reader 2's segmentations. In other words, the model agreed with Reader 1 to a similar level as Reader 2 agreed with Reader 1 for all subregions (Figure 3). The same is true for how closely the model agreed with Reader 2 relative to how closely Reader 1 agreed with Reader 2.

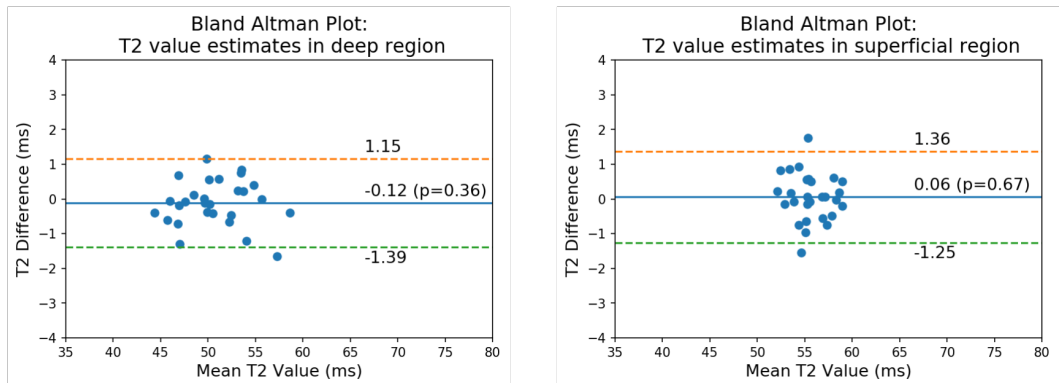

**Figure 2: Bland Altman Plots for the deep (left) and superficial (right) cartilage regions' mean T2.** Each point represents one test set subject at one time point. For example, if a point were located at (50,1), this would mean that the model's estimate of T2 value was 1 ms higher than Reader 1's estimate for a subject with a mean T2 value of 50 ms. Dotted lines represent 1.96 standard deviations above and below the mean difference between the model's and Reader 1's estimates. The model's estimates of mean T2 value in the deep and superficial cartilage regions did not display significant bias relative to Reader 1. This suggests that the model did not systematically underestimate or overestimate the T2 value in either region. Plots for smaller subregions can be found in Supplementary Material 7.

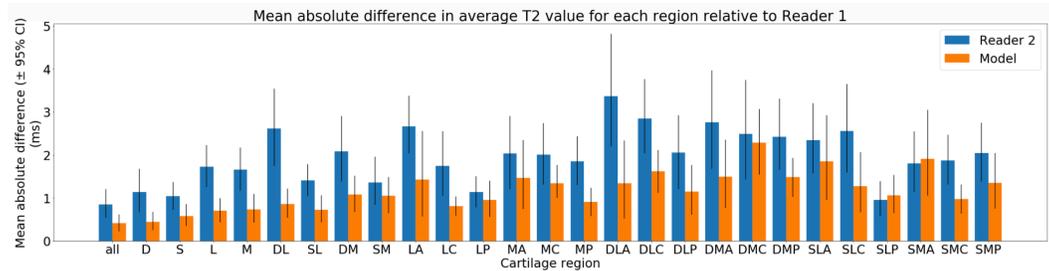

**Figure 3: Mean error of subregional average T2 estimates relative to Reader 1.** Blue bars indicate the difference between estimates derived from Reader 2's segmentations and those of Reader 1. Orange bars indicate the difference between estimates derived from the model's segmentations and those of Reader 1. Error bars indicate the 95% confidence interval. Errors for Reader 2 and the model are comparable for all subregions. In other words, the model agreed with Reader 1 to a similar level as Reader 2 agreed with Reader 1 when estimating subregional T2 values. This suggests that the model could be used to make detailed, expert-quality assessments of cartilage health in cross-sectional studies. *Subregion abbreviation key*: **all**: full cartilage plate, **D**: deep 50% of cartilage plate, **S**: superficial 50% of cartilage plate, **L**: lateral, **M**: medial, **A**: anterior, **C**: central, **P**: posterior.

### 3.4 Comparison of longitudinal subregional T2 change

The model's estimates of four-year change in subregional average T2 values were significantly correlated with those of Reader 1 for all subregions ($p < 0.01$) except the superficial lateral anterior subregion ($p = 0.08$). Spearman correlations and the MAE in the model's estimates of subregional average T2 change are shown in (Figure 4, Right). The lateral anterior subregion of one subject was found to have a large, full-thickness lesion that prevented a meaningful analysis of average T2 change for this subregion of this subject and so it was excluded. The other cartilage subregions of this subject were included in the analysis. Eleven of the 12 smallest subregions' model estimates did not have significant bias ($p > 0.05$) while the deep medial central subregion had a bias of -2.63 ms (Supplementary Material 7: Bland Altman Plots).

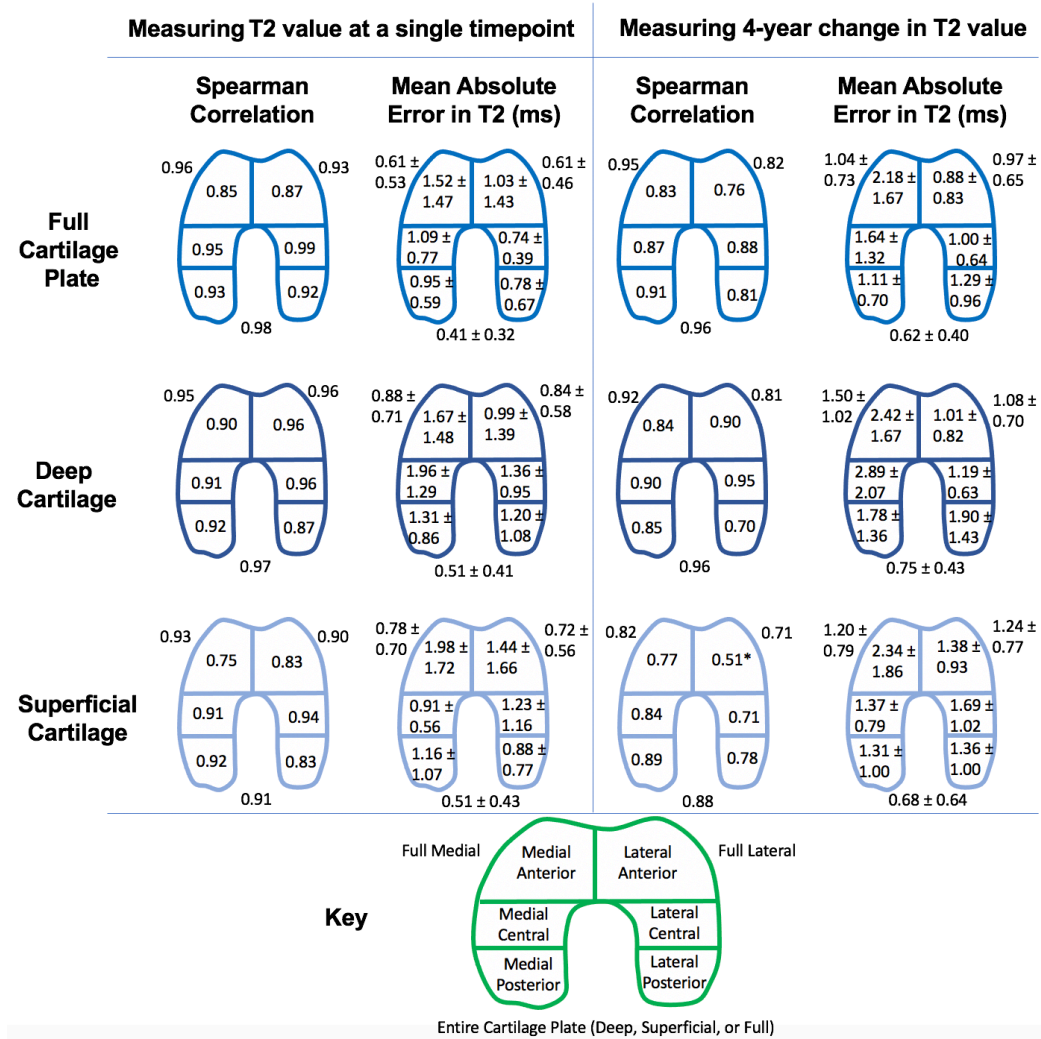

**Figure 4: Model's agreement with gold standard in its estimates of subregional mean T2 values (Left) and in its estimates of four-year change in subregional mean T2 values (Right).**
Absolute errors are reported as mean ± standard deviation. All correlations are significant ($p < 0.01$) except for the four-year change in the superficial lateral anterior subregion (p = 0.08, denoted with *). Estimates of subregional T2 values all have a MAE less than 2 ms and estimates of four-year change all have a MAE less than 3 ms. This is comparable to inter-reader agreement.

### 3.5 Comparison of focal T2 elevation

The percentage of cartilage area affected by significantly increasing T2 clusters estimated using model segmentations correlated with those of Reader 1 with a Spearman correlation of 0.78 ($p<0.01$). The MAE in the estimates was $(1.7 \pm 1.1)$% of the cartilage plate. The average percentage of cartilage area affected was 8.83% when calculated using Reader 1 segmentations and 8.67% when calculated using model segmentations.

## 4 Discussion

We have developed a fast, fully-automated, end-to-end femoral cartilage segmentation model that agrees with an expert as closely as two experts agree with one another, both in terms of segmentation masks and downstream assessments of cartilage health. The model operates directly on MESE images, eliminating the need to capture a morphological scan in addition to the MESE image to measure T2 values. We have made our model, along with code for replicating the results of this manuscript, publicly available at https://github.com/kathoma/AutomaticKneeMRISegmentation.

Several limitations of our work should be noted. The model's performance was assessed using semi-automated segmentation refined by an expert as the gold standard, which varies depending on the reader. Also, our test subset that was segmented by Reader 2 was only 14 images. However, the small standard deviation in the Dice scores and mean absolute differences in subregional T2 values observed across those 14 images suggest that these results may be representative of inter-expert agreement more broadly. The model was trained only on subjects without radiographic osteoarthritis, so its performance on images featuring gross morphological disease requires additional investigation. However, a key potential benefit of quantitative MRI is the opportunity to detect OA before it is visible on X-rays. Because analyzed subjects had knee pain and were at risk of developing OA according to OAI criteria, they made an ideal cohort for assessing this use case. Reader 2 did not use the same semi-automated segmentation approach as our gold standard, but instead manually segmented the images. Both semi-automated and manual segmentation are commonly used for knee cartilage in MESE images, so comparisons between these approaches provide valuable context for assessing the reproducibility of our model relative to different standard practices. The fact that our model agrees with Reader 1 segmentations more closely than Reader 2 segmentations suggests that our model has learned to replicate the nuanced, fine details of Reader 1 that other readers may disagree with. However, this is an issue that currently limits comparisons between any two studies in the body of knee cartilage literature that use different readers to segment their images. By making our model publicly available, we introduce the potential to enhance comparisons between future studies that leverage our model in their work.

Other works have aimed to automatically segment T2-weighted knee MRIs using atlas-based approaches, shape models, and deep learning. One work developed a deep learning model for fat-suppressed T2-weighted fast spin-echo images[37]. They report a femoral cartilage segmentation Dice score of 0.81 ± 0.04, similar to our model's 0.85 ± 0.03. Although they used T2-weighted images, these images do not provide quantitative T2 maps like the MESE images we used.

In a prior work that used all MESE images from the OAI baseline cohort, images were segmented via multiple steps of non-rigid registrations with an atlas image[33]. Relative to our approach, this is significantly more time-consuming. In estimates of the T2 value in anatomical regions of the femoral cartilage, they reported a MAE of 2.16 ms for the lateral femur (Pearson correlation $R$ = 0.82) and 1.73 ms for the medial femur (Pearson correlation $R$ = 0.75). In contrast, our model has a MAE of 0.61 ms for the lateral femur (Pearson correlation $R$ = 0.96) and 0.61 ms for the medial femur (Pearson correlation $R$ = 0.96) (Supplementary Material 5: Pearson correlation coefficients). They report an average bias of -1.2 while our bias for the medial femur is -0.1 and for the lateral femur is 0.06 (Supplementary Material 7: Bland Altman Plots). They did not report the demographics or OA severities of the subjects in their test set, which may affect these comparisons. However, no subjects in our test set had a four-year change in lateral femoral cartilage T2 less than our model's MAE of 0.61 ms, but 12 of 14 subjects had a four-year change larger than their model's MAE 2.16 ms. Similarly, 1 of 14 subjects had a four-year change in medial femoral cartilage T2 less than our model's MAE of 0.61 ms, but 12 of 14 subjects had a four-year change larger than their model's MAE 1.73 ms. This suggests that our model provides clinically significant improvements in the ability to detect longitudinal change.

A third work combined a deep learning model with a 3D simplex deformable model to segment MESE images[31]. Their simplex deformable modeling step required 56% of their algorithm's total segmentation computation time, which our method avoids. Their approach used T2 maps as input to their segmentation system, in contrast to our use of the second echo image. By eliminating the need to calculate T2 values for all non-cartilage voxels, our approach also reduces the computational time of this step by approximately two orders of magnitude. These time-saving features come with no cost in model performance. From their reported results, it can be derived that their model had a MAE of 1.4 ms in estimating the average T2 value of the full femoral cartilage plate. This is higher than our MAE of 0.41 ± 0.32 ms. It can also be derived that their Jaccard index was 0.75 ± 0.06, comparable to our model's 0.74 ± 0.04.

It is important to note that the images and researchers performing reference segmentations differed between our work and each of these prior works, for which models are not publicly available. This precludes ideal comparisons of accuracy and performance between models. We have therefore made our model and segmentation data publicly available to facilitate future comparisons.

Our model's performance may be near the limit of reproducibility for measuring regional T2 values. In an assessment of scan-rescan reproducibility of T2 values in healthy controls imaged twice on the same day, it was found that the medial and lateral femoral cartilage estimates' root mean squared coefficient of variation (RMS-CV) were between 4.0-4.5%[38]. In contrast, our model's RMS-CV in estimating T2 values were 1.46% for the lateral femoral cartilage and 1.48% for the medial femoral cartilage (Supplementary Material 6). In another study, inter-reader RMS-CV for whole knee cartilage mean T2 was 1.57%[24], while our model's is 1.0%. Our model reproduces the measurements of an expert on individual images more closely than a single experts' measurements of healthy controls agree with one another in the scan-rescan context and with a similar level of agreement as two experts reading the same image.

Although Dice scores and average T2 values for large cartilage regions are useful for assessing model performance, we go further. First, we assessed model performance using smaller subregions of the femoral cartilage. It is more difficult for two readers (or a reader and model) to have strong agreement on the average T2 value in smaller subregions because the mean is calculated over fewer voxels. Yet it is important to assess how well models agree with experts on the small subregions that result from splitting the cartilage along medial-lateral, superficial-deep, and anterior-central-posterior boundaries because this is frequently how T2 values are tracked in research[39–41]. Second, we assessed how well the model is able to capture changes in T2 value over time for each subregion. T2 values are known to vary significantly across individuals, including healthy controls[33]. The efficacy of treatments is therefore often assessed by the change in patients' T2 value over time. We measure how well our model tracks longitudinal T2 change in two ways: (1) calculating changes in subregional mean T2 values over time, and (2) tracking the extent of focal areas of T2 worsening. While agreement between expert readers regarding these longitudinal metrics is not assessed here and is scarce in prior research, the error in the model's estimates of subregional T2 change over 4 years are similar to the disagreement in subregional T2 value between our two readers for single time points.

In conclusion, we present an open source, fast and fully-automated femoral cartilage segmentation model that agrees with experts as closely as experts agree with one another. This makes it possible to leverage MESE-based findings in large-scale studies and has the potential to unlock new lines of inquiry on the earliest stage of OA.

**Acknowledgments**
This work was supported by NIH grant P41 EB027060 and NIH Big Data to Knowledge (BD2K) Research Grant U54EB020405.

**References**

1. Jordan JM, Helmick CG, Renner JB, et al. Prevalence of knee symptoms and radiographic and symptomatic knee osteoarthritis in African Americans and Caucasians: the Johnston County Osteoarthritis Project. *J Rheumatol*. 2007;34(1):172-180.

2. Felson DT, Naimark A, Anderson J, Kazis L, Castelli W, Meenan RF. The prevalence of knee osteoarthritis in the elderly. The Framingham Osteoarthritis Study. *Arthritis Rheum*. 1987;30(8):914-918.

3. Bagge E, Bjelle A, Valkenburg HA, Svanborg A. Prevalence of radiographic osteoarthritis in

two elderly European populations. *Rheumatol Int*. 1992;12(1):33-38.

4. Kellgren JH, Lawrence JS. Radiological Assessment of Osteo-Arthrosis. *Annals of the Rheumatic Diseases*. 1957;16(4):494-502. doi:10.1136/ard.16.4.494

5. Guermazi A, Roemer FW, Burstein D, Hayashi D. Why radiography should no longer be considered a surrogate outcome measure for longitudinal assessment of cartilage in knee osteoarthritis. *Arthritis Res Ther*. 2011;13(6):247.

6. Andriacchi TP, Mündermann A, Smith RL, Alexander EJ, Dyrby CO, Koo S. A framework for the in vivo pathomechanics of osteoarthritis at the knee. *Ann Biomed Eng*. 2004;32(3):447-457.

7. Eckstein F, Burstein D, Link TM. Quantitative MRI of cartilage and bone: degenerative changes in osteoarthritis. *NMR Biomed*. 2006;19(7):822-854.

8. Mosher TJ, Dardzinski BJ. Cartilage MRI T2 relaxation time mapping: overview and applications. *Semin Musculoskelet Radiol*. 2004;8(4):355-368.

9. Dunn TC, Lu Y, Jin H, Ries MD, Majumdar S. T2 relaxation time of cartilage at MR imaging: comparison with severity of knee osteoarthritis. *Radiology*. 2004;232(2):592-598.

10. Chaudhari AS, Kogan F, Pedoia V, Majumdar S, Gold GE, Hargreaves BA. Rapid Knee MRI Acquisition and Analysis Techniques for Imaging Osteoarthritis. *J Magn Reson Imaging*. Published online November 21, 2019. doi:10.1002/jmri.26991

11. Heimann T, Morrison BJ, Styner MA, Niethammer M, Warfield S. Segmentation of knee images: a grand challenge. In: *Proc. MICCAI Workshop on Medical Image Analysis for the Clinic*. ; 2010:207-214.

12. Norman B, Pedoia V, Majumdar S. Use of 2D U-Net Convolutional Neural Networks for Automated Cartilage and Meniscus Segmentation of Knee MR Imaging Data to Determine Relaxometry and Morphometry. *Radiology*. 2018;288(1):177-185. doi:10.1148/radiol.2018172322

13. Tamez-Peña JG, Farber J, González PC, Schreyer E, Schneider E, Totterman S. Unsupervised Segmentation and Quantification of Anatomical Knee Features: Data From the Osteoarthritis Initiative. *IEEE Transactions on Biomedical Engineering*. 2012;59(4):1177-1186.

14. Dam EB, Lillholm M, Marques J, Nielsen M. Automatic segmentation of high- and low-field knee MRIs using knee image quantification with data from the osteoarthritis initiative. *J Med Imaging (Bellingham)*. 2015;2(2):024001.

15. Raj A, Vishwanathan S, Ajani B, Krishnan K, Agarwal H. Automatic knee cartilage segmentation using fully volumetric convolutional neural networks for evaluation of osteoarthritis. In: *2018 IEEE 15th International Symposium on Biomedical Imaging (ISBI 2018)*. ; 2018:851-854.

16. Wirth W, Eckstein F, Kemnitz J, et al. Accuracy and longitudinal reproducibility of quantitative femorotibial cartilage measures derived from automated U-Net-based segmentation of two different MRI contrasts: data from the osteoarthritis initiative healthy reference cohort. *MAGMA*. Published online October 6, 2020. doi:10.1007/s10334-020-00889-7

17. Fürst D, Wirth W, Chaudhari A, Eckstein F. Layer-specific analysis of femorotibial cartilage t2 relaxation time based on registration of segmented double echo steady state (dess) to multi-

echo-spin-echo (mese) images. *MAGMA*. Published online May 26, 2020. doi:10.1007/s10334-020-00852-6

18. Urish KL, Keffalas MG, Durkin JR, Miller DJ, Chu CR, Mosher TJ. T2 texture index of cartilage can predict early symptomatic OA progression: data from the osteoarthritis initiative. *Osteoarthritis Cartilage*. 2013;21(10):1550-1557.

19. Zhong H, Miller DJ, Urish KL. T2 map signal variation predicts symptomatic osteoarthritis progression: data from the Osteoarthritis Initiative. *Skeletal Radiol*. 2016;45(7):909-913.

20. Peterfy CG, Schneider E, Nevitt M. The osteoarthritis initiative: report on the design rationale for the magnetic resonance imaging protocol for the knee. *Osteoarthritis Cartilage*. 2008;16(12):1433-1441.

21. Wirth W, Maschek S, Eckstein F. Sex- and age-dependence of region- and layer-specific knee cartilage composition (spin-spin-relaxation time) in healthy reference subjects. *Ann Anat*. 2017;210:1-8.

22. Hannila I, Lammentausta E, Tervonen O, Nieminen MT. The repeatability of T2 relaxation time measurement of human knee articular cartilage. *MAGMA*. 2015;28(6):547-553.

23. Jungmann PM, Kraus MS, Alizai H, et al. Association of metabolic risk factors with cartilage degradation assessed by T2 relaxation time at the knee: data from the osteoarthritis initiative. *Arthritis Care Res* . 2013;65(12):1942-1950.

24. Stehling C, Baum T, Mueller-Hoecker C, et al. A novel fast knee cartilage segmentation technique for T2 measurements at MR imaging--data from the Osteoarthritis Initiative. *Osteoarthritis Cartilage*. 2011;19(8):984-989.

25. Jordan CD, McWalter EJ, Monu UD, et al. Variability of CubeQuant T1$\rho$, quantitative DESS T2, and cones sodium MRI in knee cartilage. *Osteoarthritis Cartilage*. 2014;22(10):1559-1567.

26. Duryea J, Iranpour-Boroujeni T, Collins JE, et al. Local area cartilage segmentation: a semiautomated novel method of measuring cartilage loss in knee osteoarthritis. *Arthritis Care Res* . 2014;66(10):1560-1565.

27. Schaefer LF, Sury M, Yin M, et al. Quantitative measurement of medial femoral knee cartilage volume - analysis of the OA Biomarkers Consortium FNIH Study cohort. *Osteoarthritis Cartilage*. 2017;25(7):1107-1113.

28. Williams AA, Titchenal MR, Do BH, Guha A, Chu CR. MRI UTE-T2* shows high incidence of cartilage subsurface matrix changes 2 years after ACL reconstruction. *J Orthop Res*. 2019;37(2):370-377.

29. Chaudhari AS, Black MS, Eijgenraam S, et al. Five-minute knee MRI for simultaneous morphometry and T2 relaxometry of cartilage and meniscus and for semiquantitative radiological assessment using double-echo in steady-state at 3T. *J Magn Reson Imaging*. 2018;47(5):1328-1341.

30. Eijgenraam SM, Chaudhari AS, Reijman M, et al. Time-saving opportunities in knee osteoarthritis: T2 mapping and structural imaging of the knee using a single 5-min MRI scan. *Eur Radiol*. 2020;30(4):2231-2240.

31. Liu F, Zhou Z, Jang H, Samsonov A, Zhao G, Kijowski R. Deep convolutional neural network and 3D deformable approach for tissue segmentation in musculoskeletal magnetic resonance

imaging. *Magn Reson Med*. 2018;79(4):2379-2391.

32. Desai AD, Gold GE, Hargreaves BA, Chaudhari AS. Technical Considerations for Semantic Segmentation in MRI using Convolutional Neural Networks. *arXiv [eessIV]*. Published online February 5, 2019. http://arxiv.org/abs/1902.01977

33. Pedoia V, Lee J, Norman B, Link TM, Majumdar S. Diagnosing osteoarthritis from T2 maps using deep learning: an analysis of the entire Osteoarthritis Initiative baseline cohort. *Osteoarthritis Cartilage*. 2019;27(7):1002-1010.

34. Ronneberger O, Fischer P, Brox T. U-Net: Convolutional Networks for Biomedical Image Segmentation. *arXiv [csCV]*. Published online May 18, 2015. http://arxiv.org/abs/1505.04597

35. Smith HE, Mosher TJ, Dardzinski BJ, et al. Spatial variation in cartilage T2 of the knee. *J Magn Reson Imaging*. 2001;14(1):50-55.

36. Monu UD, Jordan CD, Samuelson BL, Hargreaves BA, Gold GE, McWalter EJ. Cluster analysis of quantitative MRI T2 and T1ρ relaxation times of cartilage identifies differences between healthy and ACL-injured individuals at 3T. *Osteoarthritis Cartilage*. 2017;25(4):513-520.

37. Liu F, Zhou Z, Samsonov A, et al. Deep Learning Approach for Evaluating Knee MR Images: Achieving High Diagnostic Performance for Cartilage Lesion Detection. *Radiology*. 2018;289(1):160-169.

38. Li X, Pedoia V, Kumar D, et al. Cartilage T1ρ and T2 relaxation times: longitudinal reproducibility and variations using different coils, MR systems and sites. *Osteoarthritis Cartilage*. 2015;23(12):2214-2223.

39. Bittersohl B, Hosalkar HS, Sondern M, et al. Spectrum of T2* values in knee joint cartilage at 3 T: a cross-sectional analysis in asymptomatic young adult volunteers. *Skeletal Radiol*. 2014;43(4):443-452.

40. Liu F, Chaudhary R, Hurley SA, et al. Rapid multicomponent T2 analysis of the articular cartilage of the human knee joint at 3.0T. *J Magn Reson Imaging*. 2014;39(5):1191-1197.

41. Souza RB, Kumar D, Calixto N, et al. Response of knee cartilage T1rho and T2 relaxation times to in vivo mechanical loading in individuals with and without knee osteoarthritis. *Osteoarthritis Cartilage*. 2014;22(10):1367-1376.

# Supplementary Material

## SM.1 Osteoarthritis Initiative patient identification numbers

Our training set was composed of the baseline and Year 4 MRIs for the following subjects:

9626869, 9584428, 9064631, 9590002, 9504627, 9575193, 9186589, 9137709, 9480533, 9501860, 9804177, 9753871, 9528886, 9307098, 9244698, 9072758, 9742066, 9606218, 9688844, 9879321, 9815509, 9157251, 9877295, 9461271, 9807132, 9433383, 9272118, 9649967, 9010952, 9691359, 9448099, 9512848, 9669143, 9003126, 9951122, 9298954, 9065433, 9086868, 9678997, 9829221, 9833782, 9855117, 9301641, 9668982, 9172581, 9481542, 9037697, 9608737, 9484468, 9958526, 9685238, 9250986, 9652149, 9692604, 9250222, 9668142, 9011949, 9950320, 9124024, 9923837, 9048979, 9962045, 9129477, 9624954, 9446305, 9144547, 9322375, 9673712, 9738052, 9710415, 9580736, 9300212, 9657071, 9891264, 9608502, 9781589, 9453107, 9326364, 9211054, 9185786, 9642695, 9104043, 9460225, 9088396, 9973569, 9152525, 9503361, 9038962, 9543679, 9329346, 9641381, 9816138, 9895750, 9612143, 9269655, 9208055, 9609444, 9312487, 9299361, 9268509, 9848855, 9106510, 9561824, 9432327, 9559860, 9095839, 9398079, 9326076, 9295223, 9907012, 9964826, 9637393, 9461826, 9115742, 9162906

Our validation set was composed of the baseline and Year 4 MRIs for the following subjects:

9617005, 9424368, 9081879, 9676793, 9363262, 9996851, 9492880, 9244001, 9588488, 9537090, 9264649, 9715626, 9601162, 9263153

Our test set was composed of the baseline and Year 4 MRIs for the following subjects:

9543086, 9123289, 9260036, 9435250, 9909311, 9518827, 9013634, 9245760, 9458093, 9405107, 9120358, 9279874, 9376146, 9529761

The test subset that was segmented by Reader 2 was composed of the following Patient-Year pairs:

(9435250, baseline), (9013634, Year 4), (9909311, Year 4), (9279874, baseline), (9260036, Year 4), (9543086, baseline), (9123289, Year 4), (9518827, Year 4), (9245760, baseline), (9458093, Year 4), (9405107, Year 4), (9120358, Year 4), (9376146, Year 4), (9529761, Year 4)

## SM.2 Loss function

Loss functions measure how well a neural network's predictions agree with the ground truth. As a model is trained, it iteratively adjusts the patterns that it uses to make predictions in order to minimize the loss as measured by a given loss function. Loss functions must be differentiable to enable this iterative adjustment. The Dice score is a commonly used metric for assessing the agreement between two medical image segmentations. It is the harmonic mean between precision and recall, so in this context, it provides a single measure of the model's ability to detect cartilage where it is present while also avoiding labeling voxels as cartilage that do not contain cartilage. Our goal was to create a model that produces segmentations with as high of a Dice score as possible in relation to Reader 1's segmentations. However, the original formulation of the Dice score is not differentiable so we used a modified soft Dice score that is differentiable. This enabled us to train a model that directly optimizes our metric of interest.

$$DSC = \frac{2 * (True\ Cartilage\ Voxels * Predicted\ Cartilage\ Voxels)}{True\ Cartilage\ Voxels + Predicted\ Cartilage\ Voxels}$$

$$Soft\ DSC = \frac{2 * (Probability\ that\ model\ assigns\ to\ true\ cartilage\ voxels)}{Probability\ that\ model\ assigns\ to\ true\ cartilage\ voxels + True\ Cartilage\ Voxels}$$

**Figure SM1: Dice Score Formulation.** (Top) The original formulation of the dice score. This was used for the calculation of final results. (Bottom) The differentiable soft dice score formulation. This was used while training the neural network model.

### SM.3 Training hyperparameters

All models were trained using Adam optimization with $\beta_1=0.99$ and $\beta_2=0.995$. The models differed in terms of batch size, learning rate, learning rate decay, dropout rate, and the use of batch normalization. A range of value combinations were evaluated in order to find the best combination of values for these parameters. Each model was evaluated with the validation set images and the model that performed best was selected as the final model and was then evaluated with the test set.

### SM.4 Hardware and software

Models were trained and evaluated using one NVIDIA Tesla K80 GPU (Santa Clara, Calif). All modeling and analyses were done using the Python programming language version 3.5 and Keras version 2.1.5.

### SM.5 Pearson correlation coefficients

Table SM1. Pearson correlation coefficients for model estimates of average T2 value in each region relative to Reader 1 on full test set ($p < 0.0001$ for all correlations)

| Cartilage Region | Pearson correlation coefficient |
|---|---|
| All | 0.979 |
| Deep | 0.984 |
| Superficial | 0.943 |
| Lateral | 0.964 |
| Medial | 0.962 |
| Deep lateral | 0.971 |
| Superficial lateral | 0.915 |
| Deep medial | 0.961 |
| Superficial medial | 0.925 |
| Lateral anterior | 0.892 |
| Lateral central | 0.984 |
| Lateral posterior | 0.942 |
| Medial anterior | 0.876 |

*Automatic knee MRI cartilage assessment*| | |
|---|---|
| Medial central | 0.952 |
| Medial posterior | 0.963 |
| Deep lateral anterior | 0.949 |
| Deep lateral central | 0.958 |
| Deep lateral posterior | 0.914 |
| Deep medial anterior | 0.914 |
| Deep medial central | 0.922 |
| Deep medial posterior | 0.947 |
| Superficial lateral anterior | 0.806 |
| Superficial lateral central | 0.929 |
| Superficial lateral posterior | 0.911 |
| Superficial medial anterior | 0.780 |
| Superficial medial central | 0.951 |
| Superficial medial posterior | 0.932 |

**SM.6  Root mean squared coefficient of variation (RMS-CV) for model estimates to average T2 value in each region relative to Reader 1**

Table SM2. RMS-CV for model estimates to average T2 value in each region relative to Reader 1 for full test set

| **Cartilage Region** | **RMS-CV** |
|---|---|
| All | 0.97% |
| Deep | 1.30% |
| Superficial | 1.20% |
| Lateral | 1.46% |
| Medial | 1.48% |
| Deep lateral | 2.02% |
| Superficial lateral | 1.68% |
| Deep medial | 2.21% |
| Superficial medial | 1.84% |
| Lateral anterior | 3.27% |
| Lateral central | 1.66% |
| Lateral posterior | 1.96% |
| Medial anterior | 3.65% |
| Medial central | 2.48% |
| Medial posterior | 2.11% |
| Deep lateral anterior | 3.33% |
| Deep lateral central | 3.45% |
| Deep lateral posterior | 3.14% |
| Deep medial anterior | 4.09% |
| Deep medial central | 4.64% |
| Deep medial posterior | 3.09% |
| Superficial lateral anterior | 3.90% |
| Superficial lateral central | 3.20% |
| Superficial lateral posterior | 2.17% |
| Superficial medial anterior | 4.25% |
| Superficial medial central | 1.87% |
| Superficial medial posterior | 2.85% |

## SM.7 Bland Altman Plots

Bland Altman plots for the model's estimation of average T2 value in each region relative to Reader 1 for the full test set. The solid blue line represents the average difference (i.e. bias), the dashed orange line represents +1.96 standard deviations, and the dashed green line represents -1.96 standard deviations. Ten of the 12 smallest regions' model estimates did not have significant bias ($p > 0.05$) while the superficial lateral anterior region had a bias of 0.87 ms and the deep medial central region had a bias of -1.02 ms. This suggests that the model has little-to-no systematic error is it's estimates of T2 value, even for small subregions of cartilage. *Region abbreviation key*: **all**: full cartilage plate, **D**: deep 50% of cartilage plate, **S**: superficial 50% of cartilage plate, **L**: lateral, **M**: medial, **A**: anterior, **C**: central, **P**: posterior.

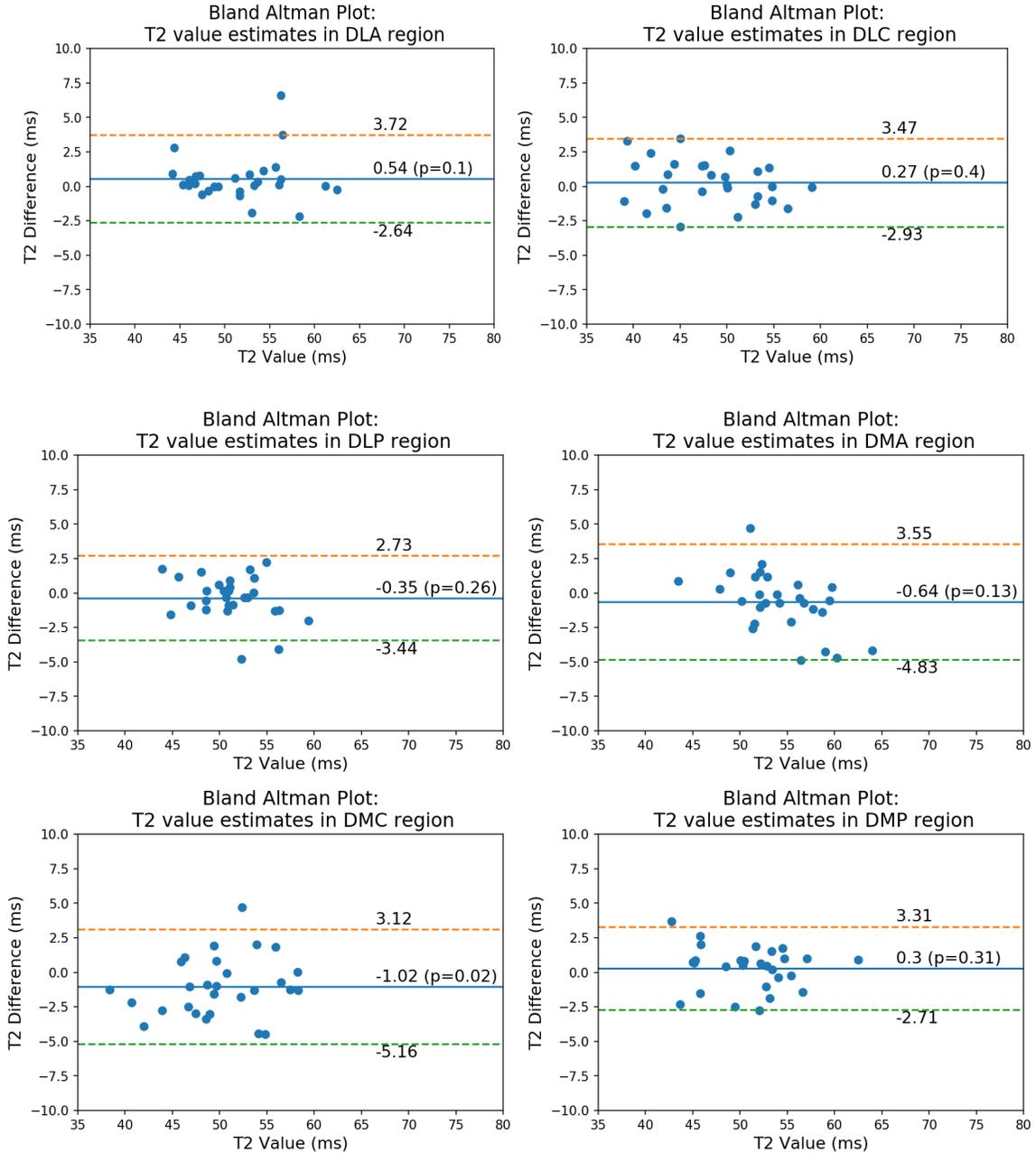

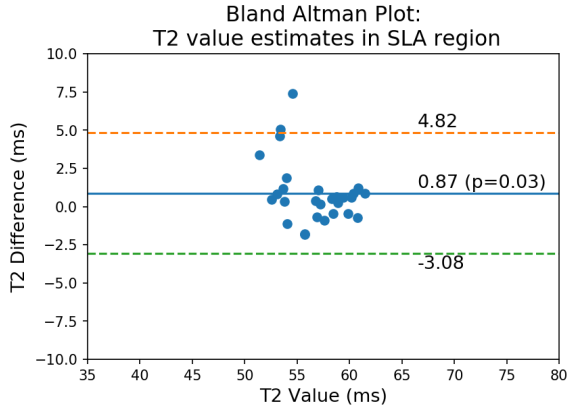
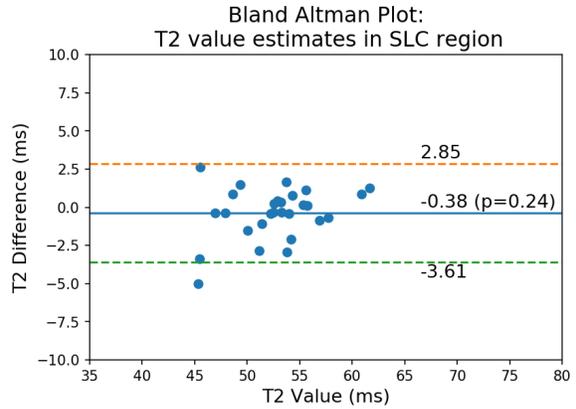
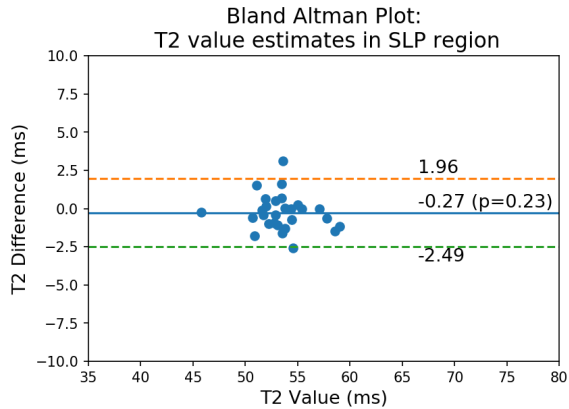
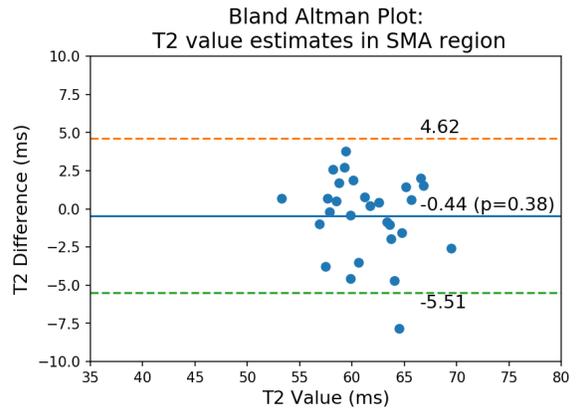
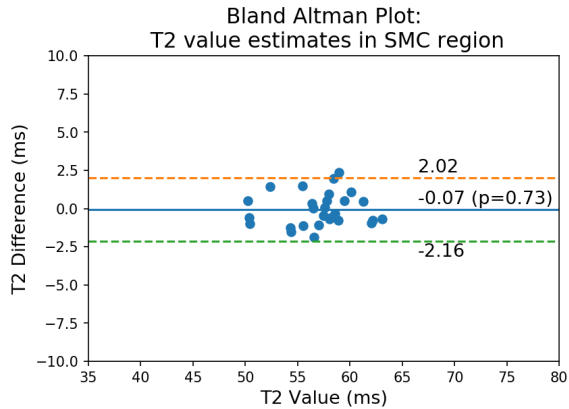
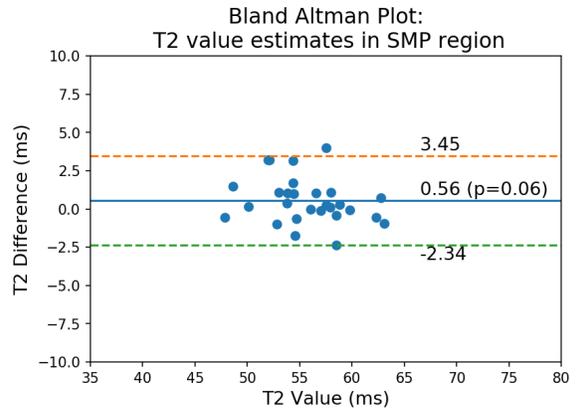

Bland Altman plots for the model's estimation of T2 change in each region relative to Reader 1 for the full test set. The solid blue line represents the average difference (i.e. bias), the dashed orange line represents +1.96 standard deviations, and the dashed green line represents -1.96 standard deviations. Eleven of the 12 smallest regions' model estimates did not have significant bias (p > 0.05) while the deep medial central region had a bias of -2.63 ms. This suggests that the model has little-to-no systematic error is it's estimates of longitudinal T2 change, even for small subregions of cartilage. *Region abbreviation key*: **all**: full cartilage plate, **D**: deep 50% of cartilage plate, **S**: superficial 50% of cartilage plate, **L**: lateral, **M**: medial, **A**: anterior, **C**: central, **P**: posterior.

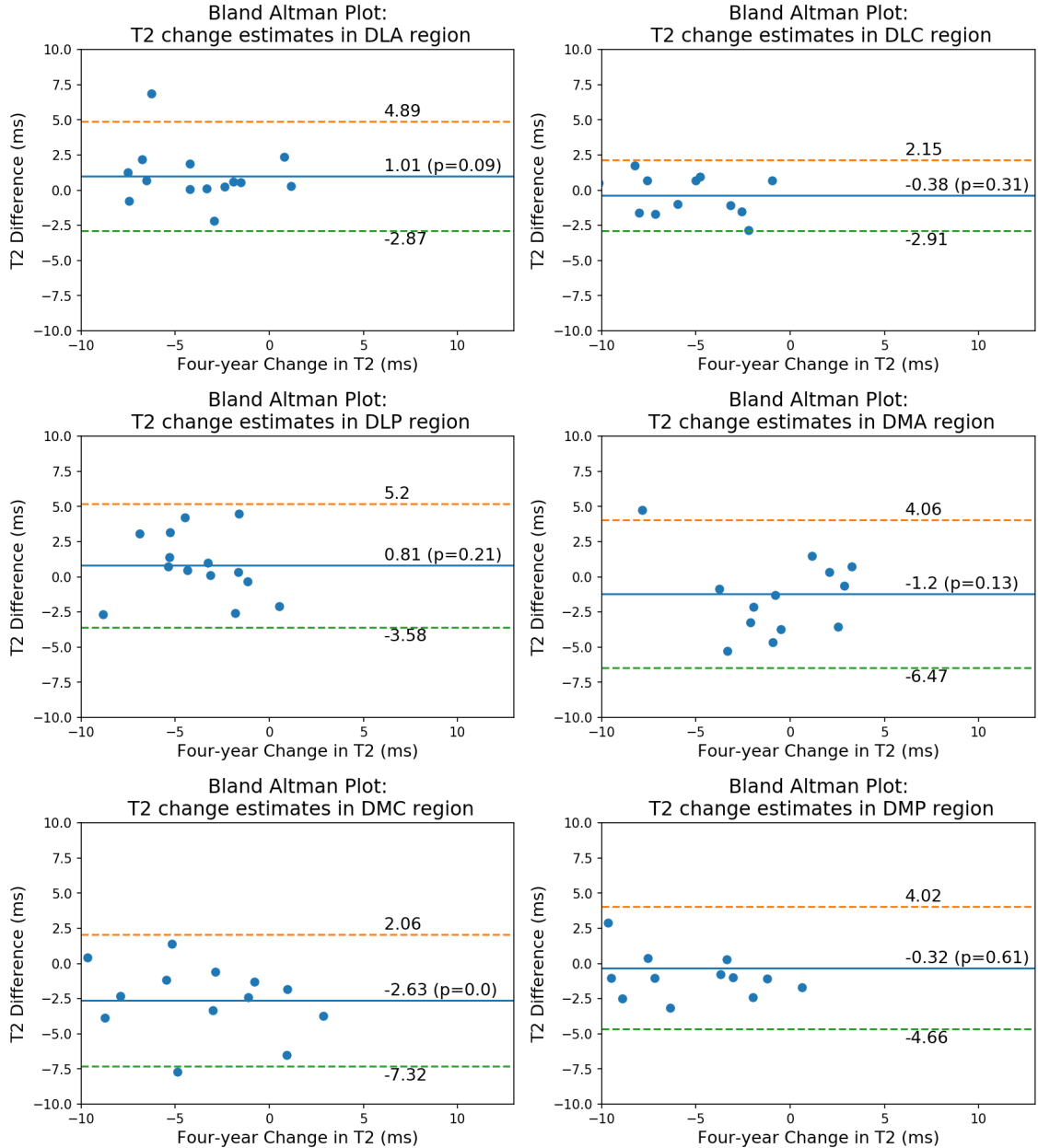

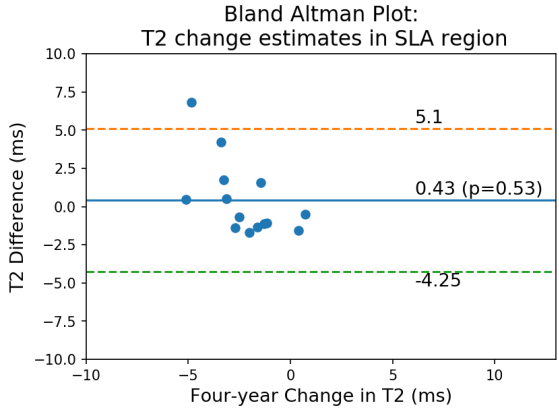
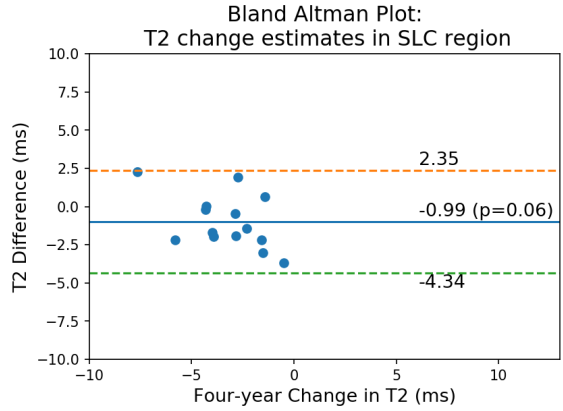
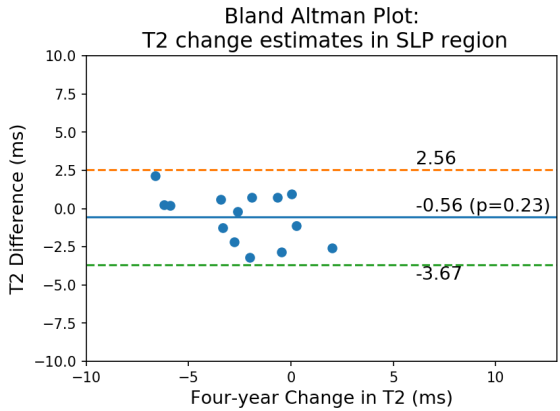
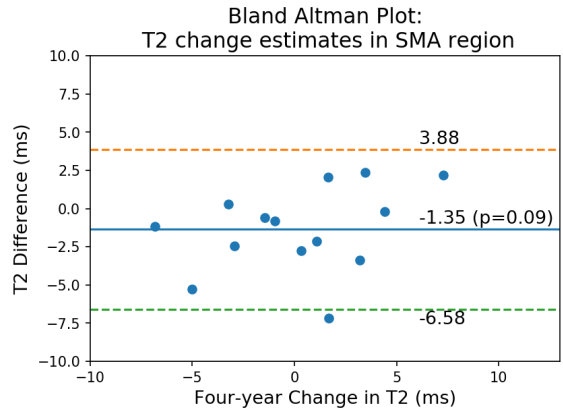
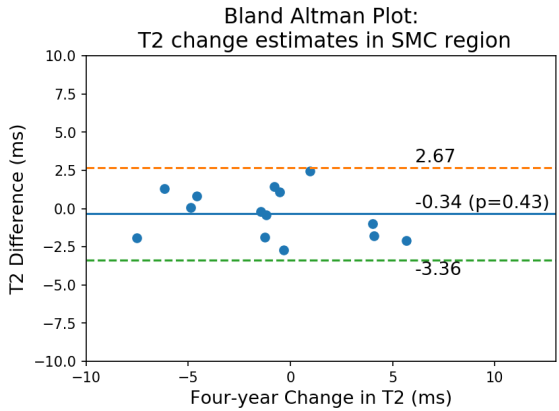
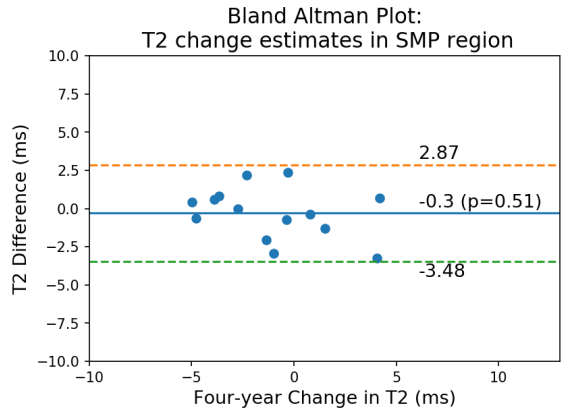

.